\pdfoutput=1

\documentclass[11pt]{article}

\usepackage[]{acl}

\usepackage{times}
\usepackage{latexsym}
\usepackage{amssymb}
\usepackage{hyperref}
\usepackage{graphicx}
\usepackage{amsmath}
\usepackage{multirow}
\usepackage{array}
\usepackage{algorithm}
\usepackage{algpseudocode}
\usepackage{pifont}
\usepackage{color}
\usepackage{colortbl}
\usepackage[T1]{fontenc}

\usepackage[utf8]{inputenc}

\usepackage{microtype}

\title{Unifying Structure and Language Semantic for Efficient Contrastive Knowledge Graph Completion with Structured Entity Anchors}

\author{Sang-Hyun Je, Wontae Choi, Kwangjin Oh\\
  Naver Corp. Seongnam, Republic of Korea. \\
  \texttt{\{sanghyun.je,choi.wontae,kj.oh\}@navercorp.com} \\}

\begin{document}
\maketitle

\begin{abstract}

The goal of knowledge graph completion (KGC) is to predict missing links in a KG using trained facts that are already known. 
In recent, pre-trained language model (PLM) based methods that utilize both textual and structural information are emerging, but their performances lag behind state-of-the-art (SOTA) structure-based methods or some methods lose their inductive inference capabilities in the process of fusing structure embedding to text encoder.
In this paper, we propose a novel method to effectively unify structure information and language semantics without losing the power of inductive reasoning.
We adopt entity anchors and these anchors and textual description of KG elements are fed together into the PLM-based encoder to learn unified representations.
In addition, the proposed method utilizes additional random negative samples which can be reused in the each mini-batch during contrastive learning to learn a generalized entity representations.
We verify the effectiveness of the our proposed method through various experiments and analysis. The experimental results on standard benchmark widely used in link prediction task show that the proposed model outperforms existing the SOTA KGC models. Especially, our method show the largest performance improvement on FB15K-237, which is competitive to the SOTA of structure-based KGC methods.

\end{abstract}

\section{Introduction}

Freebase \cite{bollacker2008freebase}, YAGO \cite{suchanek2007yago}, and WordNet \cite{miller1995wordnet} are popular large-scale knowledge graphs (KGs) that contain human knowledge information as form of directed graph structure.
\begin{figure}[h]
    \centering
    \includegraphics[width=\columnwidth]{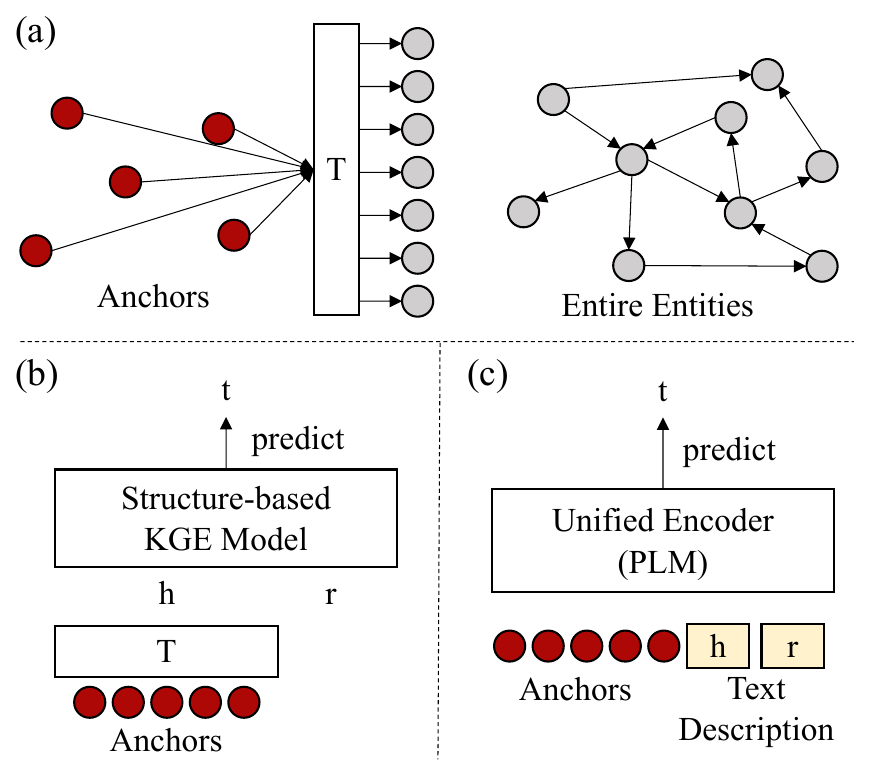}
    \caption{(a) Transformation from the fixed sized anchors to full entity embeddings. (b) The anchors are trained via structure-based KGE method. (c) The anchors and descriptions of $h$, $r$ are fed into unified model to predict tail. Since all entities are encoded by their description and the fixed anchors, the model can represent unseen entities.}
    \label{fig:simple_overview}
\end{figure}
In general, KGs are represented as a set of triple facts. A \texttt{fact} (or \texttt{triple}) consists of two entities and a relation, which is usually expressed in the form of \texttt{(head entity, relation, tail entity)}, which is also denoted as \texttt{(h, r, t)}. The most of the KGs are inherently incomplete, therefore, KGC techniques are used to infer the missing links from the known facts of KGs to enrich KGs and these methods are helpful to various downstream tasks that utilize knowledge information.

There are two main streams of KGC methods: structure based and textual description based methods. Traditional structure based methods use only the structure information of KG to learn the representation of entities and relations. These methods, such as TransE \cite{bordes2013translating}, DistMult \cite{yang2014embedding}, ComplEx \cite{trouillon2016complex}, RotatE \cite{sun2019rotate}, and TukER \cite{balavzevic2019tucker}, map each entity and relation into a low-dimensional vector. On the other hand, with development of PLM and large language model (LLM), textual description based KGC methods are emerging \cite{yao2019kg, wang2021kepler, choi2021mem, wang2022simkgc, pan2023unifying}. These kind of methods express entities and relations by encoding their textual information. The advantages of these methods are remarkable performance in KGs where language semantic is important, and possibility of inductive inference because they can encode  entities not seen in training time. However, the disadvantages of these methods are the high training cost due to LM encoder and the poor performance compared to SOTA structure-based KGC methods. Many researches propose various methods \cite{wang2021structure, shen2022joint, wang2022language, chen2023dipping} to deal with the lack of structure information, but some methods \cite{wang2022language, chen2023dipping}, in the process of using structure information, the inductive reasoning ability is lost which is the key advantage of textual description models.

In this paper, we propose a novel method that effectively integrates structural and textual information without losing inductive reasoning capabilities. The proposed method also has the advantage of being scalable with respect to the size of the traget KG. Inspired by the anchor \& transformation mechanism \cite{liang2020anchor, galkin2021nodepiece}, we design our model to utilize the decomposition of entity embeddings to represent every entity with a small number of anchors. These entity anchors are trained through the structure based KGC method and they are reused as input to the entity encoder, allowing them to be effectively fused with the language semantics. Since the entity anchors can be used exactly the same for unseen entities, they do not impede the model to making inductive inferences.

In addition, we find a limitation of in-batch contrastive learning in our experiments. SimKGC \cite{wang2022simkgc} enables efficient contrastive learning through in-batch negative sampling, and our proposed model also follows the approach to learn representations of entities and relations. However, we argue that in-batch negative sampling may introduce a bias that the negative samples follow the training data distribution. We address this issue by adding uniform random negative sampling strategy and we can take remarkable performance gains.

We evaluate the proposed method on FB15K-237, WN18RR, and Wikidata5M, the most widely used benchmark datasets in the KGC task. The proposed method not only outperforms SOTA text-based methods on FB15K-237 and Wikidata5M, but also shows competitive results to SOTA structure-based methods. The result of the proposed method are comparable to the performance of SOTA on WN18RR. Moreover, we conduct various analysis to identify the contribution of each component of the model.

\section{Related Work}

\subsection{Structure Based Graph Embedding Methods} \label{Structured Based Graph Embedding Methods}
There have been numerous attempts to represent entities and their associated relations as low-dimensional embeddings. Structure-based methods, such as TransE \cite{bordes2013translating}, aim to fit triples in a knowledge graph by translating the head entity's vector \texttt{h} to the tail entity's vector \texttt{t} through the relation vector \texttt{r}. DistMult \cite{yang2014embedding} employs a streamlined bi-linear interaction process that incorporates relations as diagonal elements within relation-specific diagonal matrices. ComplEx \cite{trouillon2016complex} utilizes complex-valued vectors to map entities and relations, capturing intricate relational patterns through bi-linear interactions between entity vectors. However, these methods struggle to capture all relation patterns, including symmetric, anti-symmetric, inversion, and composition. RotatE \cite{sun2019rotate} efficiently infers these patterns by mapping entities and relations to a complex vector space and framing relations as rotations from source to target entities. Despite the success of these methods in mapping embeddings, they may not perform well in an inductive setting.

\subsection{Textual Description Based Graph Embedding Methods}
Recent success with LLMs has led to a new wave of knowledge graph embedding (KGE) methods. KG-BERT \cite{yao2019kg} considers entities, relations, and triples as textual sequences, effectively reframing knowledge graph completion as a sequence classification challenge. MEM-KGC \cite{choi2021mem} leverages the principles of a masked language model to enhance the KGC task. StAR \cite{wang2021structure} introduces a hybrid model that combines textual encoding and graph embedding paradigms to harness the synergistic advantages of both. Its Siamese-style textual encoder extends the applicability of graph embeddings to entities and relations that were previously unseen. While knowledge graph completion methods using text descriptions have been criticized for having long inference times with lower performance, SimKGC \cite{wang2022simkgc} addresses this problem with efficient contrastive learning. It adopts a bi-encoder structure for training with InfoNCE \cite{oord2018representation} loss and incorporates three types of negatives: in-batch, pre-batch, and self-negatives.

\subsection{Vocabulary Reduction}
In the context of knowledge graph representation, conventional approaches involve assigning a unique embedding vector to each entity, leading to linear memory consumption growth and high computational costs when dealing with real-world knowledge graphs. NodePiece \cite{galkin2021nodepiece} is an anchor-based method that learns a fixed-size entity vocabulary, similar to subword tokenization in natural language processing (NLP). By constructing a vocabulary of subword/sub-entity units from anchor nodes in a graph with known relation types, NodePiece facilitates encoding and embedding for any entity, including previously unseen entities during training. Anchor \& Transform (ANT) \cite{liang2020anchor} learns a small set of anchor embeddings and a sparse transformation matrix. Notably, ANT represents the embeddings of entities as sparse linear combinations of these anchor embeddings, weighted according to the values within the transformation matrix. This approach is scalable and flexible, providing users with the ability to define anchors and impose additional constraints on transformations, potentially tailoring the model to specific domains or tasks.

\begin{figure*}[h]
    \centering
    \includegraphics[width=\textwidth]{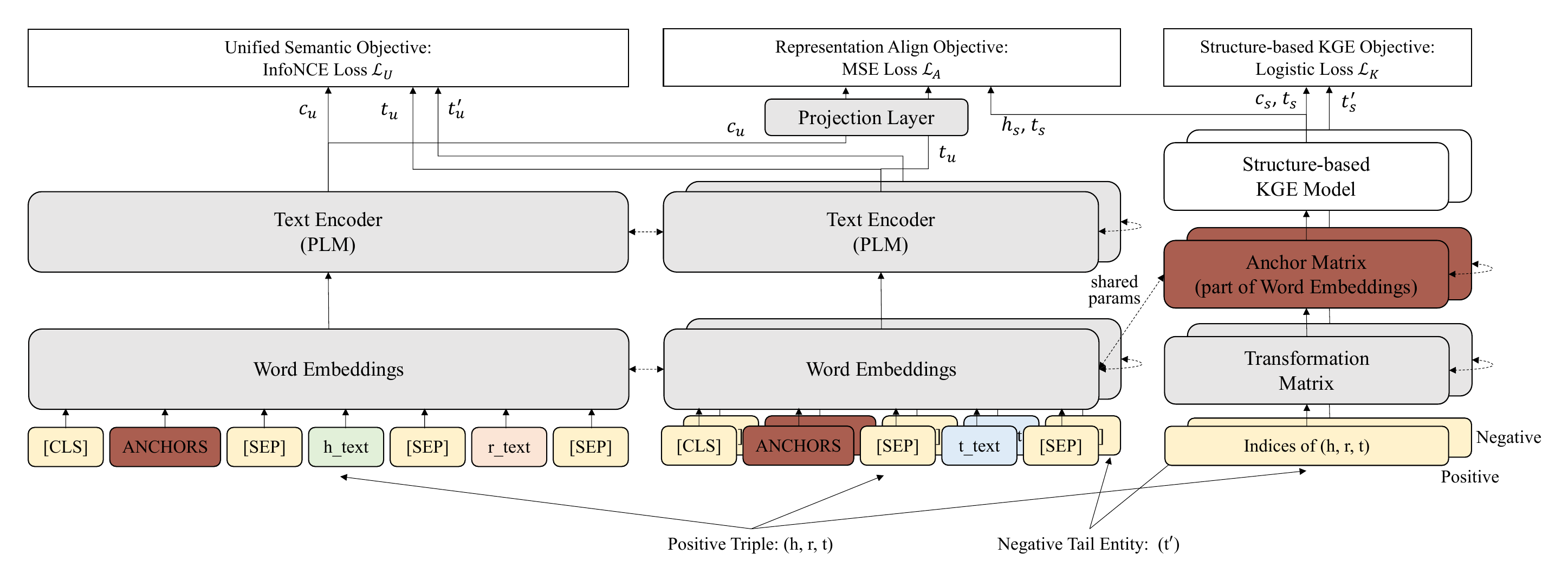}
    \caption{An overview of the SEA-KGC model architecture.}
    \label{fig:large_overview}
\end{figure*}

\section{Method}

In this section, we introduce our novel proposed unified representation learning with \textbf{S}tructured \textbf{E}ntity \textbf{A}nchors for \textbf{KGC} (\textbf{SEA-KGC}) model. We elaborate the entire architecture of the proposed model in section \ref{model_architecture}. After then, we describe the ways of initializing entity anchors used by the model in section \ref{initializing_structured_entity_anchors}. In section \ref{negative_sampling},  we explain the uniform negative sampling to complement traditional in-batch negative sampling. The following sections show the process of training and inference of the proposed model.

\subsection{Model Architecture} \label{model_architecture}

The SEA-KGC consists of two main part. The first part has PLM layer. It uses both entity anchors and textual descriptions of triple elements as input to learn a unified representation. The other part uses only entity anchors to learn the structure information of triples. This part has entity anchor matrix, transformation matrix and KGE scoring model. The losses from each part are combined into one objective, and both parts are trained simultaneously. However, we only use the first part that has PLM to get unified representations for predicting entities. In the following sections, we describe in detail the training and inference of the proposed model.

\subsubsection{Efficient Contrastive learning for Unified Representation}

The first module is the most important part of the proposed method. It utilizes the both of structure entity anchors and textual description of triple elements to generate unified representations. In this part, we encode entities and relations using PLM. Here we use BERT \cite{devlin2018bert} as  the PLM layer. All entities and relations are described into their name or description text. Each structure entity anchor is treated as special token which is part of word embeddings. The sequence of entity anchors, entity, and relation are fed together into the PLM as input.

Given a triple, we split it into two parts \texttt{(h, r)} and \texttt{(t)} and encode them separately. This approach is more efficient than encoding the entire triple at once during training and inference, as shown in several previous methods \cite{wang2021structure, wang2022simkgc}. To obtain the  representation of head entity and relation, we  make an input sequence by concatenating the entity anchors, text of head entity and text of relation with \texttt{'[SEP]'} as a separator. The combined input sequence is fed to the PLM encoder. By concatenating both structure entity anchors and entity text as a single input sequence, we can expect an integration of text and structure information. After then, we take the hidden states of the last layer of the PLM. We apply mean pooling to get the context embedding $c_u$. 

The tail entity representation $t_u$ is created by encoding the entity anchors and text of tail entity with \texttt{'[SEP]'} token. We use a single PLM that shares the same parameters for encoding head entity, relation and tail entity. The matching score between embedding $c_u$ and $t_u$ is defined by cosine similarity.

\begin{equation}
\label{eqn:cosine-sim}
    cos(c_u, t_u) = \frac{c_u \cdot t_u }{|| c_u ||\  ||t_u||}
\end{equation}

Through the optimization, we make the score between the positive pair $c_u$, $t_u$ high and the negative pair $c_u$, $t_u'$ low.

\subsubsection{Structure based Knowledge Embedding Learning}

We perform additional optimization to ensure that entity anchors capture more structure information of KG. We adopt anchor \& transformation mechanism \cite{liang2020anchor} to represent every structure entity embedding as a linear combination of the entity anchors. The entity anchor matrix and the transformation matrix containing weights of each anchor are also updated through training of KGC. Therefore, the entire entity anchors are shared with each entity and each entity maintains an entity specific transformation weight vector. We optimize the structure entity embeddings using the structure based KGE method. Here, we use TransE \cite{bordes2013translating}, the most widely used KGE scoring model. Given a triple, we compute the triple score  using the following formula with structure embedding $h_s$, $r_s$, and $t_s$.

\begin{equation}
\label{eqn:kge-transe}
    KGE(h, r, t) = -||h_s + r_s -t_s||
\end{equation}

\begin{equation}
\label{eqn:anchor-transformation}
    h_s = T[h_i] \cdot A,  r_s = R[r_i], t_s = T[t_i] \cdot A
\end{equation}

Through the training, the score between the positive pair is also getting high and the negative pair is getting low.

\subsection{Initializing Structured Entity Anchors} \label{initializing_structured_entity_anchors}

Before training the SEA-KGC model, we should initialize structure entity anchors. There are two initializing methods. The first method is to initialize entity anchors randomly. In this method, we do not need any pre-processing logic. We can simply initialize structure entity anchors and transformation matrix with random vectors and train SEA-KGC with these embeddings. The second method is to initialize the entity anchors with a clustering technique. To select more meaningful anchors, we first represent each entity embedding as a vector encoded from its text description by BERT. We cluster all entities into a target number of anchors, set the centroid of each cluster as an anchor. We use K-Means clustering algorithm to find centroids. We can reproduce all entity embeddings $E$ by matrix multiplication of the transformation and anchor matrix, $T$ and $A$. Therefore, initial transformation matrix can be calculated by multiplying the inverse of entity anchor matrix to entire embedding matrix. It can be formalized as,

\begin{align}
\label{eqn:eta}
    E & = TA \\
    T & = EA^{-1}
\end{align}
where $E \in \mathbb{R}^{V \times D}$, $T \in \mathbb{R}^{V \times N}$, and $A \in \mathbb{R}^{N \times D}$. $V$, $N$, and $D$ are size of entity vocabulary, number of anchors, and size of hidden dimension, respectively. After then, we train the proposed KGC model using the initialized entity anchors and transformation matrix.

\subsection{Negative Sampling} \label{negative_sampling}

We basically follow in-batch contrastive learning to train our model. This learning approach reuses the samples in the mini-batch as negative samples for each other to allow efficient learning as possible. However, we find a problem with in-batch negative sampling during our experiments: the in-batch negatives follow the entity occurrence distribution in the training dataset (i.e., entities that occur more frequently in training are used as negatives more often, and entities that barely occur are used as negatives less often). This problem may be an obstacle to learning a better representation, as some triples are repeatedly given only trivial negatives. To address this issue, we add randomness to negatives. We keep the in-batch negative samples $\mathcal{N}_{IB}$, but sample uniform random entities from the set of train entities, whose size is equal to the batch size, and use them as additional negatives $\mathcal{N}_{UR}$. Thus, the set of negatives $\mathcal{N}(h,r)$ is,

\begin{equation}
    \mathcal{N}(h,r) = \{ t' | t' \in  \mathcal{N}_{IB} \cup \mathcal{N}_{UR} \} .
\end{equation}

Additional uniform random negative samples are also reused within the mini-batch, allowing efficient learning. We observe remarkable performance improvements on various benchmark datasets by using the proposed negative sampling strategy.

\subsection{Training}
We optimize our proposed model using the total of three losses. At first, we use InfoNCE loss \cite{oord2018representation} for loss $\mathcal{L}_u$. We use cosine similarity as the scoring function for triples, which is defined by Eq \ref{eqn:cosine-sim}. The additive margin $\gamma_c$ helps the correct triple to get higher score. The temperature $\tau$ controls the importance weights of hard negatives. We make $\tau$ as a learnable parameter to learn along with the model during training \cite{wang2022simkgc}. The loss $L_u$ can be expressed as below. 

\begin{equation}
    \mathcal{L}_{u} = -\log \frac{e^{(\phi(h_u, r_ut_u)-\gamma_c)/\tau} }{e^{(\phi(h_u, r_u, t_u)-\gamma_c)/\tau}  + \sum e^{\phi (h_u, r_u, t_i')/ \tau }}
\end{equation}

Second, we use sigmoid loss with self-adversarial sampling for loss $\mathcal{L}_s$. The self-adversarial sampling loss shows good performance in many structure-based KGE models \cite{sun2019rotate}. We use TransE as KGE scoring function. The loss $\mathcal{L}_s$ make the entity anchors learn structure information directly from KG. The loss $\mathcal{L}_s$ can be formulated as, 

\begin{align}
    \mathcal{L}_s = - \log \sigma (\gamma_k - \phi(h_s, r_s, t_s )) \nonumber \\
    - \sum_{i=1}^{|\mathcal{N}|} p(h_s, r_s, t_s^{i'}) \log \sigma ( \phi(h_s, r_s, t_s^{i'}) - \gamma_k) \\
    p(h_s, r_s, t_s^{j'}) = \frac{\exp(\phi(h_s, r_s, t_s^{j'}))}{\sum_i \exp (\phi(h_s, r_s, t_s^{i'}))} .
\end{align}

Finally, we define a loss $L_a$ that aligns the representations created by anchors with the representations created by text encoding in the structure embedding space. By doing so, the two groups of embeddings can be located in similar distribution. The unified tail representations $t_u$ are passed into the projection layer $g(\cdot)$ and the representation $g(t_u)$ is now placed in the structure embedding space. We use a linear layer for $g( \cdot )$. We use MSE for the loss $\mathcal{L}_a^{mse}$ to make $g(t_u)$ and tail structure embedding $t_s$ closer.

\begin{equation}
    \mathcal{L}_a^{mse} = \frac{1}{n}\sum_{i=1}^{n}(g(t_u)-t_s)^2
\end{equation}

In addition, we define a loss $\mathcal{L}_a^{margin}$ using a margin ranking loss that makes the unified representation $g(t_u)$ closer to tail structure embedding $t_s$ and farther from head structure embedding $h_s$. We use the Euclidean distance for the distance function. The loss $\mathcal{L}_a$ is the sum of $\mathcal{L}_a^{mse}$ and $\mathcal{L}_a^{margin}$.

\begin{flalign}
    &\mathcal{L}_a^{margin} = \nonumber \\ 
    &\max(d(g(t_u), h_s) -d(g(t_u), t_s) + \gamma_m , \ 0)
\end{flalign}

\begin{equation}
    \mathcal{L}_a = \mathcal{L}_a^{mse} + \mathcal{L}_a^{margin}
\end{equation}

We simply sum the all three losses into a final loss $\mathcal{L}$. We update the parameters of the proposed model during optimizing the loss $\mathcal{L}$.

\begin{equation}
    \mathcal{L} = \mathcal{L}_u + \mathcal{L}_s + \mathcal{L}_a
\end{equation}

\subsection{Inference}
When predicting the tail entity, we use the both matching score and alignment score of the unified representation to calculate a final score. And then we find the entity which has the highest final score. At first, the unified representation matching score can be calculated as, 

\begin{equation}
    score_u = \frac{c_u \cdot t_u }{|| c_u ||\  ||t_u||} .
\end{equation}

The alignment score measures the similarity between the transformed unified representations. Since we use TransE as the structure KGE function, the similarity score is calculated by distance between $g(c_u)$ and $g(t_u)$ which can be formulated,

\begin{equation}
    score_s = - \lambda \ \cdot \ ||g(c_u) - g(t_u)||^{2}_{2} .
\end{equation}

The final score is sum of these two scores. After calculating the final score, we select the tail entity with the highest final score.

\begin{equation}
    score = score_u + score_s
\end{equation}

\begin{equation}
    arg\max_{t_i} \ score(h, r, t_i), \ t_i \in \mathcal{E}
\end{equation}

Additionally, we adopt the entity re-ranking method \cite{wang2022simkgc}. This strategy helps improve performance of text-based KGC methods by explicitly using the spatial locality of the KG. We increase the score of a candidate tail entity \texttt{t} by $\alpha$ if it is in the k-hop neighbors set of head entity \texttt{h} and decrease the score by $\beta$ if it is the same as the head entity, based on the training graph.

\begin{align}
    arg\max_{t_i} \ score(h, r, t_i) & + \alpha \mathbf{1}(t_i \in \mathcal{E}_k(h)) \nonumber \\
     & - \beta \mathbf{1}(t_i = h)
\end{align}

Since we do not use any entity-specific transformation vector, but only uses entity anchors and the text description of the entity to calculate the score, the proposed model can compute representations for any unseen entities. Therefore, the proposed method does not lose the inductive reasoning capability.

\section{Experiments}

To evaluate our model, we compare with various KGC methods on three datasets. We also perform various analyses to identify the contributions of each module, and pros and cons of SEA-KGC model. We implement our experiments using PyTorch framework and run them on 4 NVIDIA V100 GPUs.\footnote{The codes of our method will be updated soon.}

\begin{table}[h]
\centering
\resizebox{\columnwidth}{!}{
\begin{tabular}{c|ccccc}
\hline
Dataset & \#entity & \#relation & \#train & \#valid & \#test \\
\hline
FB15K237 & 14,541 & 237 & 272,115 & 17,535 & 20,466 \\
WN18RR & 40,943 & 11 & 86,835 & 3,034 & 3,134 \\
Wikidata5M(Transductive) & 4,594,485 & 822 & 20,614,279 & 5,163 & 5,163 \\
Wikidata5M(Inductive) & 4,579,609 & 822 & 20,496,514 & 6,699 & 6,894 \\
\hline
\end{tabular}}
\caption{Statistics of the benchmark datasets.}
\label{tab:data_statistics}
\end{table}

\subsection{Experiment Settings}

\subsubsection{Datasets and Evaluation Metrics}

WN18RR \cite{dettmers2018convolutional} and FB15K-237 \cite{toutanova2015observed} are subsets of WN18 \cite{bordes2013translating, miller1995wordnet} and FB15K \cite{bordes2013translating, bollacker2008freebase}, respectively. These datasets solve the problem of data leakage by removing inverse relations. WN18RR, which contains semantic relationships between words, consists of about 40K entities and 11 relations. FB15K-237, which contains open domain knowledge from Freebase, consists of about 15K entities and 237 relations. We use the textual descriptions of entities in WN18RR and FB15K-237, which are published by KG-BERT \cite{yao2019kg}. 
Wikidata5M \cite{wang2021kepler} is a more realistic sized dataset, consisting of about 5 million entities and 822 relations. There are two settings of the dataset: transductive and inductive. In the transductive setting, the entities that appear in the test set also appear in the train set. However, in the inductive setting, the train and test sets are completely separated and each has its own entity vocabularies. Therefore, the test set consists of entities that do not appear in the train set. The detailed statistics of the datasets are shown in Table \ref{tab:data_statistics}.

We use the most widely used metrics for model evaluation: mean reciprocal rank (MRR), and hits at N (Hit@N).  All evaluation metrics are computed under the filtered setting introduced in \cite{bordes2013translating}.

\begin{table*}[h]
\centering
\resizebox{\textwidth}{!}{
\begin{tabular}{ l | cccc | cccc }
\hline
\multirow{2}{*}{\textbf{Method}} &  \multicolumn{4}{c|}{\textbf{FB15K-237}} &  \multicolumn{4}{c}{\textbf{WN18RR}} \\
 & \textbf{MRR} & \textbf{Hit@1} & \textbf{Hit@3} & \textbf{Hit@10} & \textbf{MRR} & \textbf{Hit@1} & \textbf{Hit@3} & \textbf{Hit@10} \\
\hline

\multicolumn{9}{l}{\textit{Structure-based Methods}} \\
\hline

TransE \cite{bordes2013translating}$^{\dagger}$ & 27.9 & 19.8 & 37.6 & 44.1 & 24.3 & 4.3 & 44.1 & 53.2 \\
DistMult \cite{yang2014embedding}$^{\dagger}$ & 28.1 & 19.9 & 30.1 & 44.6 & 44.4 & 41.2 & 47.0 & 50.4 \\
ComplEx \cite{trouillon2016complex}$^{\dagger}$ & 27.8 & 19.4 & 29.7 & 45.0 & 44.9 & 40.9 & 46.9 & 53.0 \\
RotatE \cite{sun2019rotate}$^{\dagger}$ & 33.8 & 24.1 & 37.5 & 53.3 & 47.6 & 42.8 & 49.2 & 57.1 \\
TuckER \cite{balavzevic2019tucker}$^{\dagger}$ & 35.8 & 26.6 & 39.4 & 54.4 & 47.0 & 44.3 & 48.2 & 52.6 \\
NePTuNe \cite{sonkar2021neptune} & \underline{36.6} & \underline{27.2} & \textbf{40.4} & \underline{54.7} & 49.1 & 45.5 & 50.7 & 55.7 \\
\hline

\multicolumn{9}{l}{\textit{Text-based Methods}} \\
\hline

KG-BERT \cite{yao2019kg}$^{\dagger}$ & - & - & - & 42.0 & 21.6 & 4.1 & 30.2 & 52.4 \\
StAR \cite{wang2021structure} & 29.6 & 20.5 & 32.2 & 48.2 & 40.1 & 24.3 & 49.1 & 70.9 \\
KGT5 \cite{saxena2022sequence}$^{\ddagger}$ & 27.6 & 21.0 & - & 41.4 & 50.8 & 48.7 & - & 54.4 \\
KG-S2S \cite{chen2022knowledge} & 33.6 & 25.7 & 37.3 & 49.8 & 57.4 & 53.1 & 59.5 & 66.1 \\
SimKGC \cite{wang2022simkgc} & 33.6 & 24.9 & 36.2 & 51.1 & \textbf{66.6} & \textbf{58.7} & \textbf{71.7} & \textbf{80.0} \\
LMKE \cite{wang2022language} & 30.6 & 21.8 & 33.1 & 48.4 & 61.9 & 52.3 & 67.1 & 78.9 \\
LASS-RoBERTa \cite{shen2022joint} & - & - & - & 53.3 & - & - & - & 78.6 \\
CSProm-KG \cite{chen2023dipping} & 35.8 & 26.9 & 39.3 & 53.8 & 57.5 & 52.2 & 59.6 & 67.8 \\
\hline

SEA-KGC & \textbf{36.7} & \textbf{27.5} & \underline{40.1} & \textbf{55.3} & \underline{65.3} & \underline{57.7} & \underline{69.6} & \underline{79.5} \\
\hline
\end{tabular}
}
\caption{Main results for FB15K-237 and WN18RR. Results of [$^{\dagger}$] are taken from \cite{wang2021structure}, results of [$^{\ddagger}$] are taken from \cite{chen2023dipping}, and the other results are taken from the corresponding papers. Bold numbers represent the best and underlined numbers represent the second best.}
\label{tab:main_result}
\end{table*}

\begin{table*}[h]
\centering
\resizebox{\textwidth}{!}{
\begin{tabular}{ l | cccc | cccc }
\hline
\multirow{2}{*}{\textbf{Method}} &  \multicolumn{4}{c|}{\textbf{Wikidata5M-Transductive}} &  \multicolumn{4}{c}{\textbf{Wikidata5M-Inductive}} \\
 & \textbf{MRR} & \textbf{Hit@1} & \textbf{Hit@3} & \textbf{Hit@10} & \textbf{MRR} & \textbf{Hit@1} & \textbf{Hit@3} & \textbf{Hit@10} \\
\hline

\multicolumn{9}{l}{\textit{Structure-based Methods}} \\
\hline

TransE \cite{bordes2013translating}$^{\dagger\dagger}$ & 25.3 & 17.0 & 31.1 & 39.2 & - & - & - & - \\
RotatE \cite{sun2019rotate}$^{\dagger\dagger}$ & 29.0 & 23.4 & 32.2 & 39.0 & - & - & - & - \\
\hline

\multicolumn{9}{l}{\textit{Text-based Methods}} \\
\hline

DKRL \cite{xie2016representation}$^{\dagger\dagger}$ & 16.0 & 12.0 & 18.1 & 22.9 & 23.1 & 5.9 & 32.0 & 54.6 \\
KEPLER \cite{wang2021kepler} & 21.0 & 17.3 & 22.4 & 27.7 & 40.2 & 22.2 & 51.4 & 73.0 \\
BLP-SimpleE \cite{daza2021inductive} & - & - & - & - & 49.3 & 28.9 & 63.9 & 86.6 \\
SimKGC \cite{wang2022simkgc} & 35.8 & 31.3 & 37.6 & 44.1 & \underline{71.4} & \underline{60.9} & \underline{78.5} & \underline{91.7} \\
KGT5 \cite{saxena2022sequence} & 30.0 & 26.7 & 31.8 & 36.5 & - & - & - & - \\
CSProm-KG \cite{chen2023dipping} & \underline{38.0} & \textbf{34.3} & \underline{39.9} & \underline{44.6} & - & - & - & - \\
\hline

SEA-KGC & \textbf{38.6} & \underline{33.8} & \textbf{40.4} & \textbf{47.4} & \textbf{73.0} & \textbf{62.5} & \textbf{80.7} & \textbf{92.3} \\
\hline
\end{tabular}
}
\caption{Main results for Wikidata5M both settings. Results of [$^{\dagger\dagger}$] are taken from \cite{wang2022simkgc}. We follow the evaluation protocol used in \cite{wang2022simkgc}.}
\label{tab:main_result_2}
\end{table*}

\subsubsection{Hyperparameters}

We adopt BERT-Base \cite{devlin2018bert} for PLM layer in the SEA-KGC. The weights of PLM are fine-tuned with other parameters of SEA-KGC during training. We use AdamW \cite{loshchilov2017decoupled} with cosine learning rate decaying to optimize the experimental models. Some hyper-parameters validated in the following range, learning rate $\in$ \{1e-5, 3e-5, 5e-5\}, and $\beta$ for re-ranking $\in \{ 0.0, 0.1, 0.2 \}$. We set the other hyper-parameters with same values regardless of datasets. We fix the batch size to $512$. The number of entity anchor is $10$. The maximum input token length is $60$, included entity anchors. The temperature $\tau$ and the additive margin $\gamma_i$ for infoNCE loss are $0.05$ and $0.02$, respectively. We set $\alpha$ to $0.05$ for entity re-ranking. The margin of KGE scoring function $\gamma_k$ and margin ranking loss $\gamma_m$ are $9.0$ and $1.0$, respectively. The weight of the alignment score $\lambda$ is set to 0.01. We train SEA-KGC model $50$ epochs on WN18RR and $10$ epochs on FB15K-237. We train only $1$ epoch on Wikidata-5M(both transductive and inductive settings).

\subsection{Main Results}

We compare the performance of SEA-KGC with various text-based KGC methods and some popular structure-based methods. As shown in Table \ref{tab:main_result} and Table \ref{tab:main_result_2}, our proposed SEA-KGC model outperforms state-of-the-art methods on FB15K-237, Wikidata5M-Trans, and Wikidata5M-Ind. Our model also shows competitive performance on WN18RR. Especially, SEA-KGC outperforms not only text-based methods but also structure-based methods on FB15K-237. Many previous text-based methods have already shown better performance than structure-based methods on WN18RR. This is because the text-based methods have advantage of the better understanding of word semantics which is important in WN18RR. However, on FB15K-237 which has more dense connectivities and requires complex structural information, structure-based methods perform better than text-based method. To the best of our knowledge, the SEA-KGC is the first text-based method which outperforms SOTA structure based methods on FB15K-237 benchmark. The proposed model also shows the best performance in Wikidata5M on both transductive and inductive settings. These results show that SEA-KGC consistently achieves remarkable improvement on various KGs in difference scales.

\subsection{Ablation Studies}

\begin{table}[h]
\centering
\resizebox{\columnwidth}{!}{
\begin{tabular}{lccc}
\hline
Model & MRR & Hit@1 & Hit@10 \\
\hline
SEA-KGC & 36.7 & 27.5 & 55.3 \\
\hline
SEA-KGC w/ DistMult & 36.3 & 27.1 & 55.0 \\
SEA-KGC w/ Complex & 36.4 & 27.1 & 54.9 \\
SEA-KGC w/ RotatE & 36.4 & 27.1 & 55.2 \\
\hline
SEA-KGC w/ random initialized anchors & 36.1 & 26.6 & 55.0 \\
\hline
SEA-KGC w/o additional negatives & 32.8 & 23.6 & 51.8 \\
\hline
SEA-KGC w/o structure and alignment losses & 36.5 & 27.3 & 55.0 \\
SEA-KGC w/o alignment loss & 36.5 & 27.1 & 55.2 \\
\hline
\end{tabular}}
\caption{The results of ablation on FB15K-237.}
\label{tab:ablation_study}
\end{table}

\subsubsection{Impact of Structure KGE Functions}
\begin{figure}[h]
    \centering
    \includegraphics[width=\columnwidth]{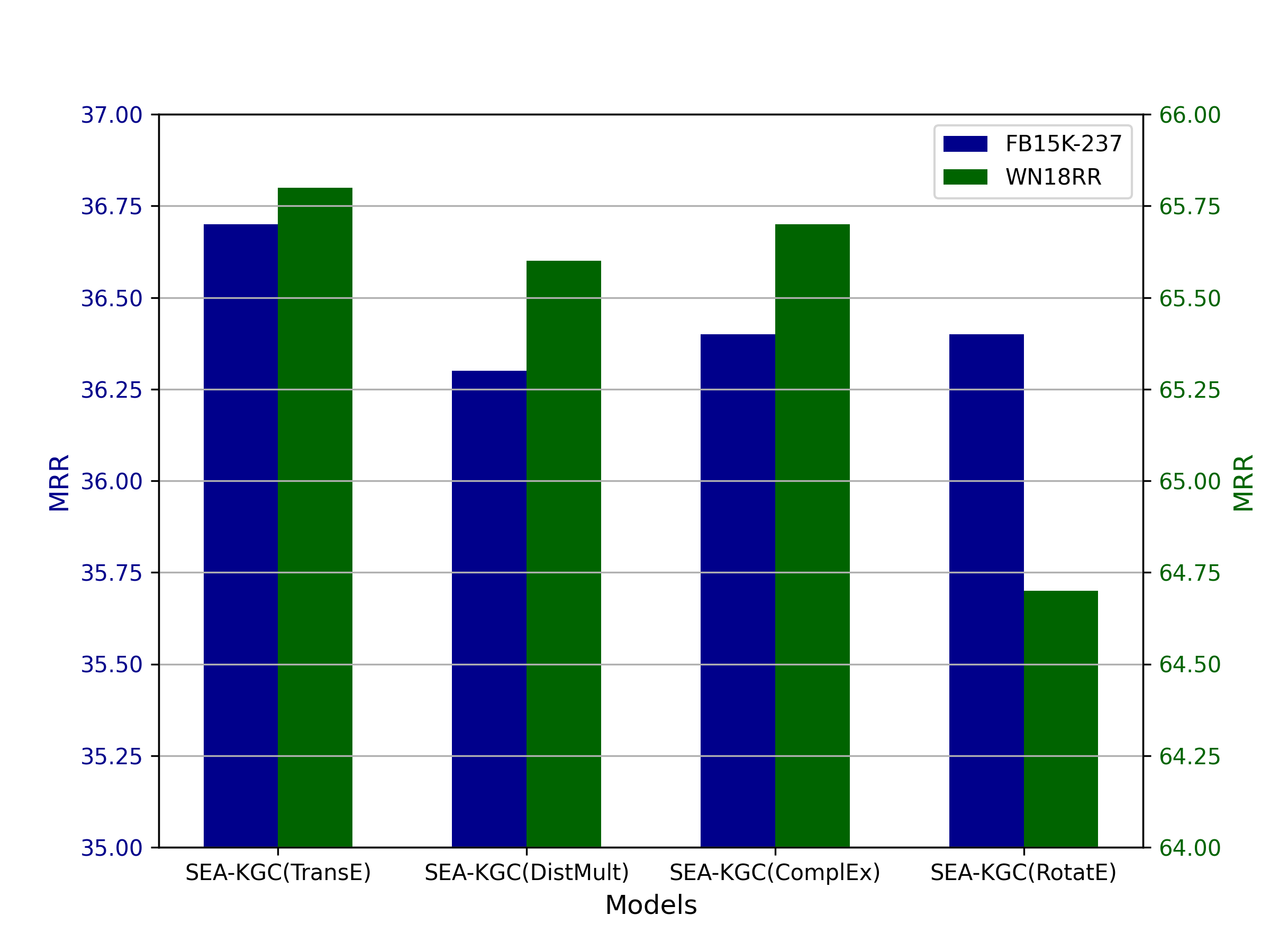}
    \caption{MRR of SEA-KGC with various KGE function on FB15K-237 and WN18RR}
    \label{fig:kge_mrrs}
\end{figure}

As presented in section \ref{Structured Based Graph Embedding Methods}, there are various structure-based knowledge graph embedding methods that can be adopted to the structural entity embeddings in SEA-KGC. To evaluate the influence of different KGE functions, we conduct experiments in which we substitute the TransE KGE scoring function with alternative KGE functions, including DistMult, ComplEx, and RotatE graph embeddings. Resultingly, it is evident that different KGE functions do not make a substantial impact on the results. In Table \ref{tab:ablation_study}, we can observe that the original SEA-KGC shows more favorable performance on FB15K-237 when compared to models employing other KGE functions, such as DistMult, ComplEx, and RotatE graph embeddings. Figure \ref{fig:kge_mrrs} also illustrates the strong MRR performance of the model using the TransE KGE function on both FB15K-237 and WN18RR datasets. While it's worth noting that a more thorough hyperparameter optimization for each KGE function may yield improved results, it appears that the SEA-KGC's unified model structure itself has a more significant impact on overall performance.

\subsubsection{Effect of Entity Anchors}

In this section, we analyze the effect of an entity anchor's initializing method on the model performance. Before KGC training, we utilize BERT to encode entire entities, cluster them, and take the centroids as anchors. As an alternative to this process, we can simply adopt random initializing for entity anchors and transformation matrices before KGC training. When applying random initializing, no additional pre-process is required before training a KGC model. However, as shown in Table \ref{tab:ablation_study}, when learning with anchors initialized to random vectors, the performance decreases slightly compared to original SEA-KGC. In particular, the performance gap is bigger in Hit@1 and MRR. Therefore, finding representative centroids through the clustering and using them as anchors is helpful in learning structured information of entities.

\begin{figure}[h]
    \centering
    \includegraphics[width=\columnwidth]{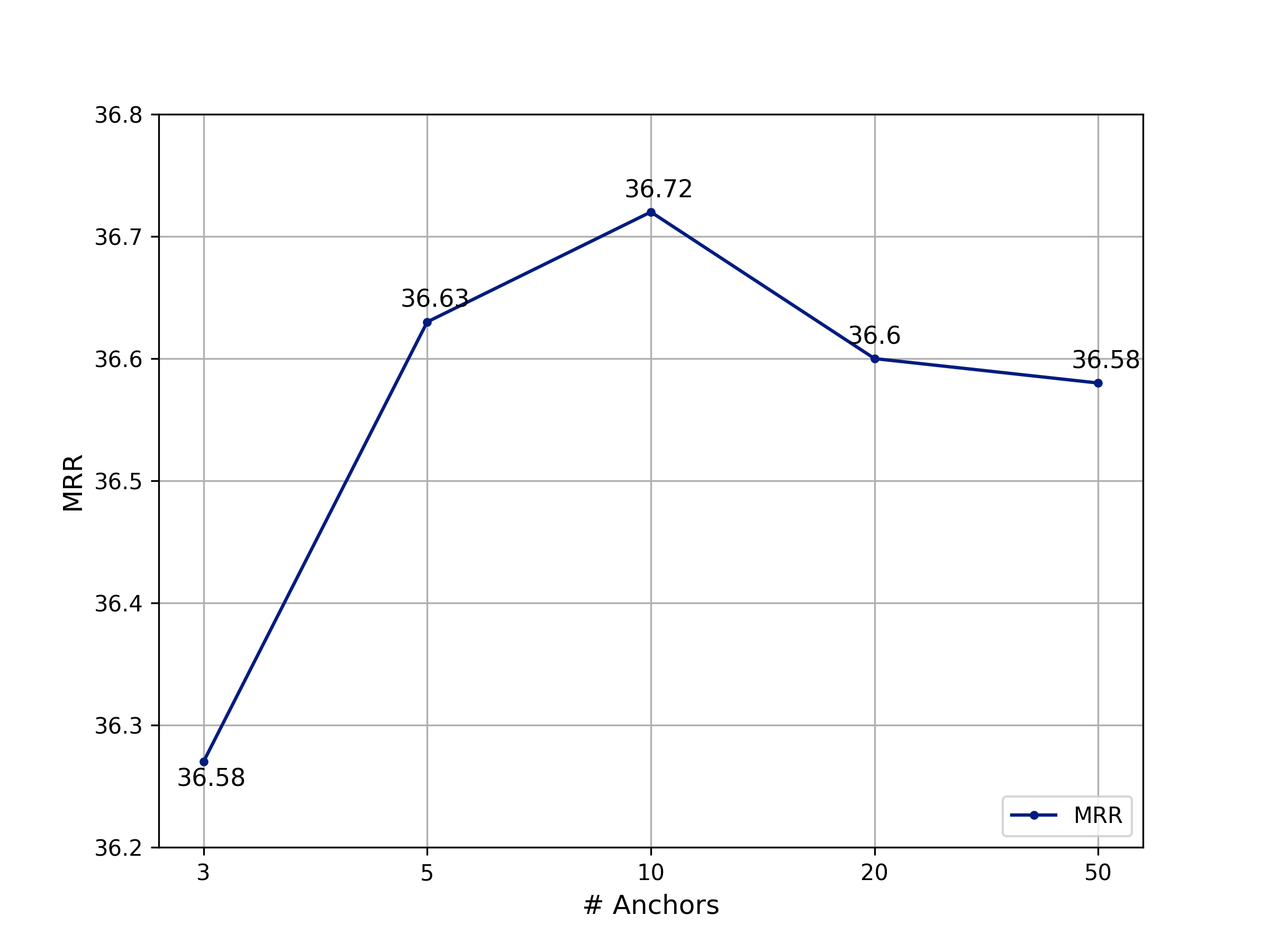}
    \caption{MRR on FB15K-237 according to the change of the number of anchors. The performance is slightly decreasing when the number of anchors is larger than 10.}
    \label{fig:mrr_n_anchors}
\end{figure}

We also perform ablation on the number of entity anchors. We compare each performance of the models when the number of entity anchor is 3, 5, 10, 20, and 50, respectively. As shown in Figure \ref{fig:mrr_n_anchors}, as the number of anchors increases, model performance also increases. However, when the number of anchors exceeds 10, performance is saturated. Rather, the performance deteriorates slightly. Moreover, since the larger the anchor size causes increase in the input sequence length for PLM, model training and inference cost more. Therefore, we should compromise between the performance and efficiency of the model with appropriate number of anchors.

\subsubsection{Effect of Negative Sampling}

We train SEA-KGC using in-batch contrastive learning with additional uniform random negatives. To check the effectiveness of additional uniform random negatives, we expreiment the training of SEA-KGC without uniform random negatives. All other settings and hyper-parameters ,except negative sampling method, are used the same as original SEA-KGC. Table \ref{tab:ablation_study} shows the results of negative sampling ablation. The performance of SEA-KGC without uniform random negatives drops by $\sim$ 4\% in all metrics compared to the original model. This result shows the most dramatic difference of all ablation experiments. Through this result, we observe that negative sampling strategy is one of the most important part in the KGC task. We also experimentally verify the limitations of in-batch negative learning and the effectiveness of uniform negative sampling.

\subsubsection{Various Losses Ablation}

The proposed model is designed to optimize total of three different losses. We perform ablation  on the loss to check the effect of each loss. First, we remove structure and alignment losses and train the SEA-KGC using only unified semantic loss. We observe that removing structure loss causes performance degradation, by 0.2 MRR and 0.3 Hit@10. This performance gap can be explained by the reason that entity anchors become meaningless tokens because they do not properly express structural information. We also observe a performance drop (i.e., 0.2 MRR and 0.1 Hit@10) when only the alignment loss is removed from SEA-KGC. This experiment verifies that explicit alignment is necessary, although unified representation has implicit alignment inherent.

\begin{table}[h]
\centering
\resizebox{\columnwidth}{!}{
\begin{tabular}{l}
\hline
\rowcolor{lightgray}
Input (head and relation): \\
\rowcolor{lightgray}
\quad (Charlotte Gainsbourg [\textit{is an Anglo-French actress and singer...}], /person/nationality) \\
\hline
Prediction (Baseline): \\
\quad P1: Normandy [\textit{is a geographical region of France...}] \\
\quad P2: Brittany [\textit{is a cultural region in the north-west of France...}] \\
\hline
Prediction (SEA-KGC): \\
\quad P1: \textbf{England [\textit{is a part of the United Kingdom...}]} \\
\quad P2: United Kingdom [\textit{The Uinted Kingdom of Great Britain and Northern Ireland...}] \\
\hline
\rowcolor{lightgray}
Input (head and relation): \\
\rowcolor{lightgray}
\quad (Theodore Roosevelt [\textit{was an American author, ... and politician, ...}], /government\_position\_held/basic\_title) \\
\hline
Prediction (Baseline): \\
\quad P1: Member of Congress [\textit{is a person who has been appointed or elected and inducted into some official body...}] \\
\quad P2: Secretary of State [\textit{is a commonly used title for a senior or mid-level post in governments...}] \\
\hline
Prediction (SEA-KGC): \\
\quad P1: \textbf{ President [\textit{is a leader of an organization, company, community, club...}]} \\
\quad P2: Secretary of State [\textit{is a commonly used title for a senior or mid-level post in governments...}] \\
\hline
\end{tabular}}
\caption{Case study on FB15K-237. Texts in brackets are description for entities. Bold text represents the ground truth entity.}
\label{tab:case_study}
\end{table}

\subsection{Case Study}

In this section, we give qualitative analysis by comparing a baseline model and our SEA-KGC. We select SimKGC \cite{wang2022simkgc} as a baseline model, because the model has simple architecture based on PLM but shows competitive performance on various benchmarks.

We show some examples to illustrate how SEA-KGC outperforms the baseline model. Table \ref{tab:case_study} shows the predictions of each model for some test examples in FB15K-237. In the first case, the baseline model ranks two different provinces in France as the predictions. This could be caused by the keyword \textit{"French"} in the description of head entity. However, SEA-KGC returns \textit{"England"} as tail entity correctly. In the second case, SEA-KGC selects correct tail entity, \textit{"President"}, as answer from many similar government positions in KG using the trained structure information while the baseline model misses the answer. This is probably because the description of the head entity contains a variety of occupation and political career information, making the baseline model confused to find the correct answer based on textual similarity. From these results, we can figure out that SEA-KGC can capture both text and structure information, while the the baseline focuses excessively on textual information.

\subsection{Error Analysis}

\begin{figure}[h]
    \centering
    \includegraphics[width=\columnwidth]{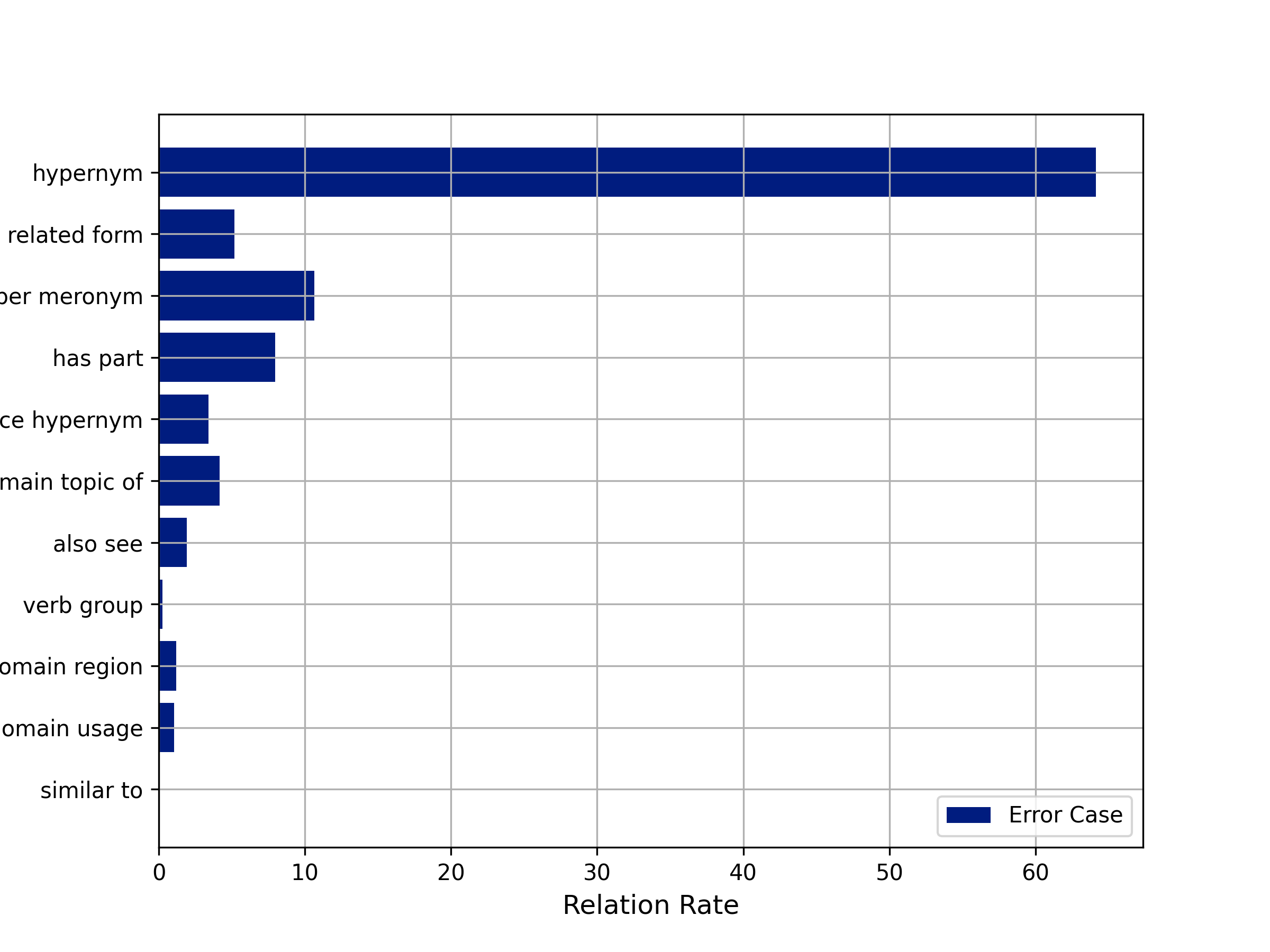}
    \caption{Error rate of each relation in test set of WN18RR. More than 60\% of the wrong predictions are linked to \textit{hypernym} relations.}
    \label{fig:error_rate_wn18rr}
\end{figure}

We perform error case analysis to confirm the limitations of SEA-KGC. We summarize the error cases that occurred in WN18RR by relation. As shown in Figure \ref{fig:error_rate_wn18rr}, the most frequent errors of the proposed method are caused by \textit{hypernym} relation. 

\begin{table}[h]
\centering
\resizebox{\columnwidth}{!}{
\begin{tabular}{l}
\hline
\rowcolor{lightgray}
Input (head and relation): \\
\rowcolor{lightgray}
\quad (Geography [\textit{study of the earth's surface...}], hypernym) \\
\hline
Prediction (Baseline): \\
\quad P1: \textbf{Earth Science [\textit{any of the sciences that deal with the earth...}]} \\
\quad P2: Subject Field [\textit{a branch of knowledge...}] \\
\hline
Prediction (SEA-KGC): \\
\quad P1: Scientific Discipline [\textit{a particular branch of scientific knowledge...}] \\
\quad P2: Subject Field [\textit{a branch of knowledge...}] \\
\hline
\rowcolor{lightgray}
Input (head and relation): \\
\rowcolor{lightgray}
\quad (Mucus [\textit{protective secretion of the mucus}], hypernym) \\
\hline
Prediction (Baseline): \\
\quad P1: \textbf{Secretion [\textit{a functionally specialised substance...}]} \\
\quad P2: Internal Secretion [\textit{the secretion of an endocrine gland...}] \\
\hline
Prediction (SEA-KGC): \\
\quad P1: Liquid Body Substance [\textit{the liquid parts of the body}] \\
\quad P1: \textbf{Secretion [\textit{a functionally specialised substance...}]} \\
\hline
\end{tabular}}
\caption{Error analysis on WN18RR. Texts in brackets are description for entities. Bold text represents the ground truth entity.}
\label{tab:error_analysis}
\end{table}

For more detailed analysis, we again compare the baseline model and SEA-KGC. We find that several error cases in which the baseline model predicts correct answer, while the SEA-KGC does not.  First, SEA-KGC tends to predict \textit{hypernym}s in a broader sense. As shown in Table \ref{tab:error_analysis}, it seems hard to say that the top-2 tail entity predictions returned by SEA-KGC are completely wrong. Second, we find that the text semantics often overlap in the description of the head entity and answer tail entity. In the first example, \textit{"earth"} in the head description and \textit{"earth"} in the tail completely overlap. In the second example, \textit{"secretion"} in the head description and \textit{"secretion"} in the tail also overlap. Therefore, in these cases, the baseline model that mainly focuses on text similarity provides better answers. On the other hand, SEA-KGC, which considers structure information, predicts a more diverse and wide hierarchical range of \textit{hypernym}s. Through error analysis, we confirm that some errors may occur in SEA-KGC due to above reasons.

\section{Conclusion}

In this paper, we propose a novel method SEA-KGC that effectively unifies text and structure information using structured entity anchors. We use entity anchors in the proposed model to  utilize structural information effectively while the model still enables inductive reasoning. Additionally, we figure out the problem with in-batch negative sampling and alleviate the problem by introducing additional uniform random negatives. In our experiments on various benchmark datasets, and SEA-KGC outperforms SOTA KGC methods on FB15K-237 and Wikidata5M and shows competitive results on WB18RR.

For future work, we plan to adopt larger language models to our method and try to develop a model that creates powerful synergy between KG information and text semantics. Another direction is negative sampling methods. We plan to study effective hard negative sampling methods that can also be used in in-batch contrastive learning.
\section{Limitations}

We utilize PLM encoder in SEA-KGC to integrate textual and structural information. As a result, the increase in computational cost of the model is much bigger than the increase in performance on  open-domain KG(e.g. FB15K-237) compared to structured based methods. SEA-KGC has a huge performance improvement on KG where word semantics is important (e.g. WN18RR), the cost cannot be ignored. Additionally, the input sequence length for PLM is fixedly increased by the number of entity anchors. It causes more computational resources for training and inference. In our experiments, a small number of anchors was sufficient, but it can be changed when training much larger KGs. In the future, we plan to overcome these issues for SEA-KGC.

\bibliography{custom}

\begin{thebibliography}{29}
\expandafter\ifx\csname natexlab\endcsname\relax\def\natexlab#1{#1}\fi

\bibitem[{Bala{\v{z}}evi{\'c} et~al.(2019)Bala{\v{z}}evi{\'c}, Allen, and Hospedales}]{balavzevic2019tucker}
Ivana Bala{\v{z}}evi{\'c}, Carl Allen, and Timothy~M Hospedales. 2019.
\newblock Tucker: Tensor factorization for knowledge graph completion.
\newblock \emph{arXiv preprint arXiv:1901.09590}.

\bibitem[{Bollacker et~al.(2008)Bollacker, Evans, Paritosh, Sturge, and Taylor}]{bollacker2008freebase}
Kurt Bollacker, Colin Evans, Praveen Paritosh, Tim Sturge, and Jamie Taylor. 2008.
\newblock Freebase: a collaboratively created graph database for structuring human knowledge.
\newblock In \emph{Proceedings of the 2008 ACM SIGMOD international conference on Management of data}, pages 1247--1250.

\bibitem[{Bordes et~al.(2013)Bordes, Usunier, Garcia-Duran, Weston, and Yakhnenko}]{bordes2013translating}
Antoine Bordes, Nicolas Usunier, Alberto Garcia-Duran, Jason Weston, and Oksana Yakhnenko. 2013.
\newblock Translating embeddings for modeling multi-relational data.
\newblock \emph{Advances in neural information processing systems}, 26.

\bibitem[{Chen et~al.(2022)Chen, Wang, Li, and Lam}]{chen2022knowledge}
Chen Chen, Yufei Wang, Bing Li, and Kwok-Yan Lam. 2022.
\newblock Knowledge is flat: A seq2seq generative framework for various knowledge graph completion.
\newblock \emph{arXiv preprint arXiv:2209.07299}.

\bibitem[{Chen et~al.(2023)Chen, Wang, Sun, Li, and Lam}]{chen2023dipping}
Chen Chen, Yufei Wang, Aixin Sun, Bing Li, and Kwok-Yan Lam. 2023.
\newblock Dipping plms sauce: Bridging structure and text for effective knowledge graph completion via conditional soft prompting.
\newblock \emph{arXiv preprint arXiv:2307.01709}.

\bibitem[{Choi et~al.(2021)Choi, Jang, and Ko}]{choi2021mem}
Bonggeun Choi, Daesik Jang, and Youngjoong Ko. 2021.
\newblock Mem-kgc: Masked entity model for knowledge graph completion with pre-trained language model.
\newblock \emph{IEEE Access}, 9:132025--132032.

\bibitem[{Daza et~al.(2021)Daza, Cochez, and Groth}]{daza2021inductive}
Daniel Daza, Michael Cochez, and Paul Groth. 2021.
\newblock Inductive entity representations from text via link prediction.
\newblock In \emph{Proceedings of the Web Conference 2021}, pages 798--808.

\bibitem[{Dettmers et~al.(2018)Dettmers, Minervini, Stenetorp, and Riedel}]{dettmers2018convolutional}
Tim Dettmers, Pasquale Minervini, Pontus Stenetorp, and Sebastian Riedel. 2018.
\newblock Convolutional 2d knowledge graph embeddings.
\newblock In \emph{Proceedings of the AAAI Conference on Artificial Intelligence}, volume~32.

\bibitem[{Devlin et~al.(2018)Devlin, Chang, Lee, and Toutanova}]{devlin2018bert}
Jacob Devlin, Ming-Wei Chang, Kenton Lee, and Kristina Toutanova. 2018.
\newblock Bert: Pre-training of deep bidirectional transformers for language understanding.
\newblock \emph{arXiv preprint arXiv:1810.04805}.

\bibitem[{Galkin et~al.(2021)Galkin, Denis, Wu, and Hamilton}]{galkin2021nodepiece}
Mikhail Galkin, Etienne Denis, Jiapeng Wu, and William~L Hamilton. 2021.
\newblock Nodepiece: Compositional and parameter-efficient representations of large knowledge graphs.
\newblock \emph{arXiv preprint arXiv:2106.12144}.

\bibitem[{Liang et~al.(2020)Liang, Zaheer, Wang, and Ahmed}]{liang2020anchor}
Paul~Pu Liang, Manzil Zaheer, Yuan Wang, and Amr Ahmed. 2020.
\newblock Anchor \& transform: Learning sparse embeddings for large vocabularies.
\newblock \emph{arXiv preprint arXiv:2003.08197}.

\bibitem[{Loshchilov and Hutter(2017)}]{loshchilov2017decoupled}
Ilya Loshchilov and Frank Hutter. 2017.
\newblock Decoupled weight decay regularization.
\newblock \emph{arXiv preprint arXiv:1711.05101}.

\bibitem[{Miller(1995)}]{miller1995wordnet}
George~A Miller. 1995.
\newblock Wordnet: a lexical database for english.
\newblock \emph{Communications of the ACM}, 38(11):39--41.

\bibitem[{Oord et~al.(2018)Oord, Li, and Vinyals}]{oord2018representation}
Aaron van~den Oord, Yazhe Li, and Oriol Vinyals. 2018.
\newblock Representation learning with contrastive predictive coding.
\newblock \emph{arXiv preprint arXiv:1807.03748}.

\bibitem[{Pan et~al.(2023)Pan, Luo, Wang, Chen, Wang, and Wu}]{pan2023unifying}
Shirui Pan, Linhao Luo, Yufei Wang, Chen Chen, Jiapu Wang, and Xindong Wu. 2023.
\newblock Unifying large language models and knowledge graphs: A roadmap.
\newblock \emph{arXiv preprint arXiv:2306.08302}.

\bibitem[{Saxena et~al.(2022)Saxena, Kochsiek, and Gemulla}]{saxena2022sequence}
Apoorv Saxena, Adrian Kochsiek, and Rainer Gemulla. 2022.
\newblock Sequence-to-sequence knowledge graph completion and question answering.
\newblock \emph{arXiv preprint arXiv:2203.10321}.

\bibitem[{Shen et~al.(2022)Shen, Wang, Gong, and Song}]{shen2022joint}
Jianhao Shen, Chenguang Wang, Linyuan Gong, and Dawn Song. 2022.
\newblock Joint language semantic and structure embedding for knowledge graph completion.
\newblock \emph{arXiv preprint arXiv:2209.08721}.

\bibitem[{Sonkar et~al.(2021)Sonkar, Katiyar, and Baraniuk}]{sonkar2021neptune}
Shashank Sonkar, Arzoo Katiyar, and Richard~G Baraniuk. 2021.
\newblock Neptune: Neural powered tucker network for knowledge graph completion.
\newblock \emph{arXiv preprint arXiv:2104.07824}.

\bibitem[{Suchanek et~al.(2007)Suchanek, Kasneci, and Weikum}]{suchanek2007yago}
Fabian~M Suchanek, Gjergji Kasneci, and Gerhard Weikum. 2007.
\newblock Yago: a core of semantic knowledge.
\newblock In \emph{Proceedings of the 16th international conference on World Wide Web}, pages 697--706.

\bibitem[{Sun et~al.(2019)Sun, Deng, Nie, and Tang}]{sun2019rotate}
Zhiqing Sun, Zhi-Hong Deng, Jian-Yun Nie, and Jian Tang. 2019.
\newblock Rotate: Knowledge graph embedding by relational rotation in complex space.
\newblock \emph{arXiv preprint arXiv:1902.10197}.

\bibitem[{Toutanova and Chen(2015)}]{toutanova2015observed}
Kristina Toutanova and Danqi Chen. 2015.
\newblock Observed versus latent features for knowledge base and text inference.
\newblock In \emph{Proceedings of the 3rd workshop on continuous vector space models and their compositionality}, pages 57--66.

\bibitem[{Trouillon et~al.(2016)Trouillon, Welbl, Riedel, Gaussier, and Bouchard}]{trouillon2016complex}
Th{\'e}o Trouillon, Johannes Welbl, Sebastian Riedel, {\'E}ric Gaussier, and Guillaume Bouchard. 2016.
\newblock Complex embeddings for simple link prediction.
\newblock In \emph{International conference on machine learning}, pages 2071--2080. PMLR.

\bibitem[{Wang et~al.(2021{\natexlab{a}})Wang, Shen, Long, Zhou, Wang, and Chang}]{wang2021structure}
Bo~Wang, Tao Shen, Guodong Long, Tianyi Zhou, Ying Wang, and Yi~Chang. 2021{\natexlab{a}}.
\newblock Structure-augmented text representation learning for efficient knowledge graph completion.
\newblock In \emph{Proceedings of the Web Conference 2021}, pages 1737--1748.

\bibitem[{Wang et~al.(2022{\natexlab{a}})Wang, Zhao, Wei, and Liu}]{wang2022simkgc}
Liang Wang, Wei Zhao, Zhuoyu Wei, and Jingming Liu. 2022{\natexlab{a}}.
\newblock Simkgc: Simple contrastive knowledge graph completion with pre-trained language models.
\newblock \emph{arXiv preprint arXiv:2203.02167}.

\bibitem[{Wang et~al.(2021{\natexlab{b}})Wang, Gao, Zhu, Zhang, Liu, Li, and Tang}]{wang2021kepler}
Xiaozhi Wang, Tianyu Gao, Zhaocheng Zhu, Zhengyan Zhang, Zhiyuan Liu, Juanzi Li, and Jian Tang. 2021{\natexlab{b}}.
\newblock Kepler: A unified model for knowledge embedding and pre-trained language representation.
\newblock \emph{Transactions of the Association for Computational Linguistics}, 9:176--194.

\bibitem[{Wang et~al.(2022{\natexlab{b}})Wang, He, Liang, and Xiao}]{wang2022language}
Xintao Wang, Qianyu He, Jiaqing Liang, and Yanghua Xiao. 2022{\natexlab{b}}.
\newblock Language models as knowledge embeddings.
\newblock \emph{arXiv preprint arXiv:2206.12617}.

\bibitem[{Xie et~al.(2016)Xie, Liu, Jia, Luan, and Sun}]{xie2016representation}
Ruobing Xie, Zhiyuan Liu, Jia Jia, Huanbo Luan, and Maosong Sun. 2016.
\newblock Representation learning of knowledge graphs with entity descriptions.
\newblock In \emph{Proceedings of the AAAI conference on artificial intelligence}, volume~30.

\bibitem[{Yang et~al.(2014)Yang, Yih, He, Gao, and Deng}]{yang2014embedding}
Bishan Yang, Wen-tau Yih, Xiaodong He, Jianfeng Gao, and Li~Deng. 2014.
\newblock Embedding entities and relations for learning and inference in knowledge bases.
\newblock \emph{arXiv preprint arXiv:1412.6575}.

\bibitem[{Yao et~al.(2019)Yao, Mao, and Luo}]{yao2019kg}
Liang Yao, Chengsheng Mao, and Yuan Luo. 2019.
\newblock Kg-bert: Bert for knowledge graph completion.
\newblock \emph{arXiv preprint arXiv:1909.03193}.

\end{thebibliography}
\bibliographystyle{acl_natbib}

\appendix 

\end{document}